\documentclass[12pt]{article}
\usepackage[cp866]{inputenc}
\usepackage[english]{babel}
\usepackage{graphicx}
\usepackage{amsmath}
\usepackage{amsfonts,amsmath,amsthm,amssymb,indentfirst}

\newtheorem{thm}{Theorem}
\newtheorem{lemma}{Lemma}

\begin{document}
\begin{center}
{\large \bf Probability of detection of an extraneous mobile object by autonomous unmanned underwater vehicles as a solution of the Buffon problem\\
Guzev M.A., Tsitsiashvili G.Sh., Osipova M.A., $^1$Sporyshev M.S.\\
Institute for Applied mathematics, Far Eastern Branch of Russian Academy Sciences,\\
690041, Vladivostok, Radio str. 7, IAM FEB RAS,\\
$^1$Far Eastern Federal University, 690950, Vladivostok, Sukhanov str. 8\\
e-mails: guzev@iam.dvo.ru, guram@iam.dvo.ru, mao1975@list.ru, coockoombra@gmail.com}
\end{center}
\vskip 1cm 
MSC2000 60J30+60J60
\vskip 1cm
{\small {\bf Abstract.} Underwater robotics addresses the problem of object detection apparatus. Offers a probabilistic formulation of the problem, which uses the reduction of the detection task to a classical task of Buffon. This formulation arises naturally in the formulation of the problem in the coordinate system associated with the apparatus. It is shown that the problem allows analysis in the presence of an asymptotic parameter, determined by the ratio of the local scan size of the apparatus to the global size of the problem under consideration.

}

\section*{Introduction}
To calculate the probability of detection of an extraneous mobile object by autonomous unmanned underwater vehicles (AUUV) is the important problem in modern underwater robotic technique. In the initial formulation, this problem relates to the tasks of the target capturing, to the differential games [1]. Various variants of this problem formulation are possible. The first option is based on the definition of the shortest trajectory, after which the robot or a group of robots will cover with their scope all the given territory [2]. The second option involves the search for a minimum number of robots that provide a guarantee of the object capture [3]. In the early works [1-3], the goal of target capturing was formulated in terms of random graphs [4], and not in terms of random sets, which led to the disappearance of the initial geometric formulation of the problem.

This drawback can be overcome by addressing to one of the key problems of the theory of geometric probability - the Buffon problem. The Buffon problem (see Figure 1) is the determination of the probability of intersection of a needle with length l, which is randomly thrown onto a plane ruled by equally spaced parallel lines a distance L apart, with any one of these lines. This problem formed the basis for stochastic geometry and was widely used in applied statistics.
\begin{center}
\includegraphics[height=1.2in]{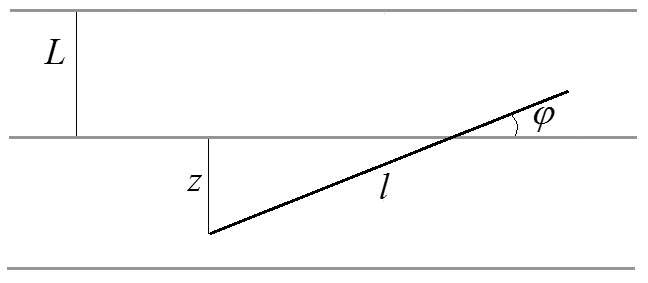}
\\{\bf Fig. 1.} 
The Buffons needle problem. 
\end{center}

The main elements of the probabilistic model in the Buffon problem are the random variables that determine the mutual position of the segment, occupied on the plane by needle, and the equally spaced horizontal lines \cite{Kendall-2}, \cite{Ambartzumian-1}. The distance $z$ from the segment lower end to the first overlying line, and the angle $\varphi<\pi$ between this line and the segment are such random variables. Knowing the distribution law of the random vector
$(z,\varphi),$ formed by these random variables, we can determine the probability of the event $P(l\sin \varphi\geq z)$ 
of the needle intersection with one of the parallel straight lines. For this event happening, it is necessary and sufficient the segment intersect with a straight line lying directly above its lower end.

As well as the authors of \cite{Monasterio}, we consider the problem of computing the probability of detection an extraneous mobile object by AUUV in terms of the Buffon problem as follows.

Let $n$ vehicles move along a circle of radius $R$ with a fixed linear velocity 
$v$ and at an equal distance each other. Each of these devices is equipped with circular radar with scanning radius of $r.$ The circular radar can be understood as a device that rotates a beam with an angular velocity large enough for the linear velocity of the beam notably exceeds the linear speed of the vehicle.

It is required to calculate the probability of detection of a mobile object by means of locators installed on the devices (see Figure 2). Under the detection, we mean here the object falling into a circle, which is scanned by any of the devices.

This work is an expanded version of the report at the VII All-Russian Scientific and Technical Conference "Technical Problems of World Ocean Exploration" \cite{Guzev}.

\section{Buffon problem in the coordinate system associated with vehicles moving round\\ a circle}
The peculiarity of the task of calculating the probability of detecting a mobile object in a stationary coordinate system, associated with an external observer, is that both the vehicle and the object are moving, and therefore some transition to the Buffon problem, in which at least parallel straight lines are fixed, is required.\begin{center}
\includegraphics[height=1.9in]{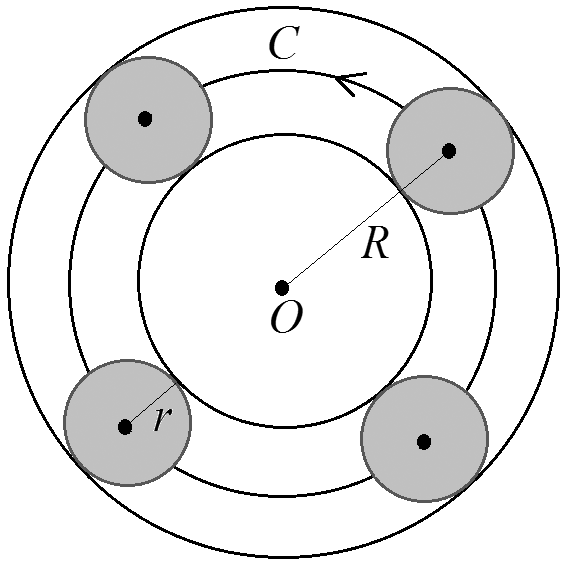}
\\{\bf Fig. 2.} 
The vehicles movement round a circle in a stationary coordinate system.
\end{center}

Such a transition is realized when the motion of the objects under study is considered in a coordinate system associated with devices rotating around a certain point $O$ round a circle $C$ of radius $R$. In this coordinate system, the radar scan circles with radius r become fixed. To be detected by the vehicle, the object trajectory should intersect one of these circles.

In order to simplify the study of this problem, we suppose that, in a stationary coordinate system, the mobile object moves along a segment connecting the starting point of its movement with the center $O$ of the circle $C$.

\begin{center}
\includegraphics[height=2.1in]{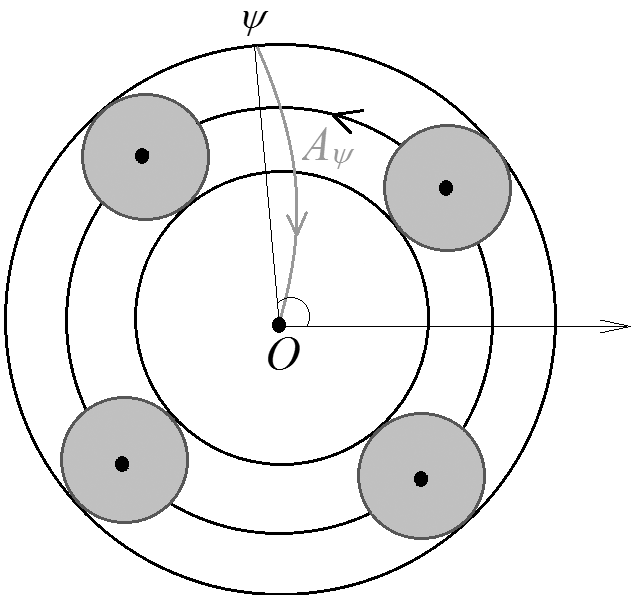}
\\{\bf Fig. 3.} 
The trajectory of the object movement of in a Cartesian coordinate system associated with vehicles moving round a circle.
\end{center}

In a rotating coordinate system, the shape of the object trajectory $A_{\psi}$ differs from the segment (see Figure 3): it is a certain curve starting at a random point $\psi$ of the circle $C_*$ with radius $R + r$ centered at the point $O.$ Thus, in the system coordinates associated with the vehicles moving round a circle, circles of radius r play the role of parallel lines, and the curves $A_{\psi}$ - the role of the segment in the Buffon problem.

In a rotating coordinate system, the shape of the trajectory $A_{\psi}$ of the object differs from the cut (see Fig. 3): it will be some curve that starts at a random point $\psi$ of the circle $C_*,$ with the radius $R+r$ and with center at $O.$ Thus, in the coordinate system, associated with the orbiting space crafts, circles of radius $r$ are playing the role of parallel lines, and curves $A_{\psi}$ -- the role of the segments in the problem of Buffon.

We further assume everywhere in this paper that the random angle $\psi$  has a uniform distribution on the segment $[0, 2\pi].$ The transition from the curve $A_{\psi^{\prime}}$ to the curve $A_{\psi^{\prime\prime}}$ is carried out by rotation of the curve $A_{\psi^{\prime}}$ around the point $O$ by the angle $\psi^{\prime\prime}-\psi^{\prime}.$ It should be noted that, on the circle $C_*,$ each circle $D_i$ with radius $r$ allocates an arc $F_i$ bounded by two curves $A_{\psi^{\prime}_i},\;A_{\psi^{\prime\prime}_i},$ that touch the circle $D_i,\; i = l, ..., n,$ from the outside (see Figure 4).

\begin{center}\vskip -0.2cm
\includegraphics[height=2.2in]{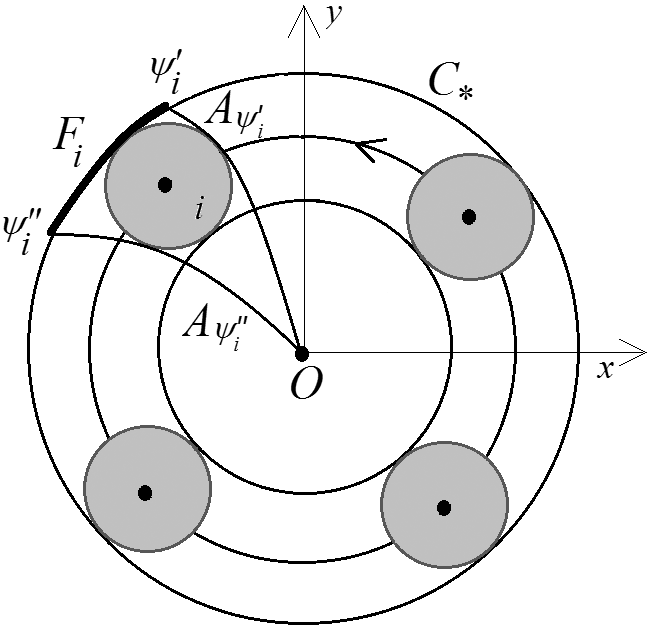}
\\{\bf Fig. 4.}  
The control area of one circle of radar scan in the Cartesian coordinate system $(x, y)$ associated with moving vehicles.
\end{center}

Thus, on the circle $C_*,$ the scan circle $D_i$ with the center at the point, where the apparatus $i$ is located, allocates the arc $F_i,\; i = 1, ..., n.$ Now we can calculate the probability of detection of a mobile object by circular view radars installed on vehicles. Let the set $\mathcal{F}$ be the union of the arcs $F_i:\;\mathcal{F}=\displaystyle \cup_{i=1}^n F_i.$ This set is the union of a finite number of disjoint arcs (maybe less than $n$). Then the total angular length of these arcs $s(\mathcal{F}),$ divided by $2\pi:\;\displaystyle {s(\mathcal{F})\over 2\pi}\;-$ is the required probability of detection a mobile object by radar locators installed on the vehicles.

Therefore, if the arcs $F_i,\; i = 1, ..., n,$ do not intersect, the probability of the object detection by the vehicles is $P=\displaystyle {ns(F_i)\over 2\pi},$ where $s (F_i)$ is the angular length of the arc $F_i.$ In the general case, $P=\displaystyle \min\left(1,\;{ns(F_i)\over 2\pi}\right).$

\section{The Buffon problem in the polar \\ coordinate system}

The task of determining the curves $A_{\psi}$ 
and finding of the arcs 
$F_i$ is solved, as a rule, numerically. 
However, at some additional assumptions, it is possible to obtain analytical formulas. Let us transit from the rectangular coordinate system 
$(x, y)$  associated with the device rotating around point 
$O$ to the polar coordinate system (see Figure 5) with the vertical coordinate
$\rho=\sqrt{x^2+y^2}$ 
and the horizontal coordinate 
$ \varphi = arctg {x\over y} . $

In a rectangular coordinate system $(x, y),$ a circle of radius $r$ centered at $(O, R)$ is described by equations 
$$x=x(\psi)=r\cos \psi,\;y=y(\psi)=R+r\sin \psi\;0\leq \psi\leq 2\pi,$$ 
giving a parametric definition of the circle: $\psi$ is a parameter. In the polar coordinate system $(\rho,\varphi)$ this circle is given parametrically by the equations 
$${\rho\over R}=\left({r^2\over R^2}\cos^2 \psi+\left(1+{r\over R}\sin \psi\right)^2\right)^{1/2},\;\varphi=arctg{{r\over R}\cos \psi\over 1+{r\over R}\sin \psi},$$
since $\psi$ remains a parameter. A typical image of circle with a radius $r$ in this polar coordinate system looks like an oval with pointed upper end (see Figure 5).
\vskip -0.1 cm 
\begin{center}
\includegraphics[height=1.7in]{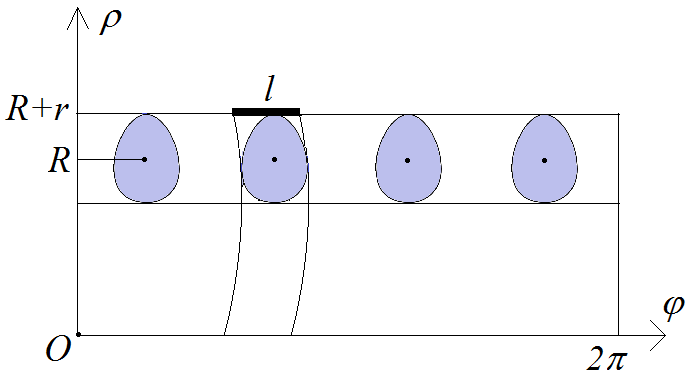}
\\{\bf Fig. 5.}
The control area of one circle of radar scan in the polar coordinate system associated with moving vehicles.
\end{center}

For small values of the parameter $r/R\ll 1,$ the last system of equations can be approximated as follows:

$${\rho\over R}\approx 1+{r\over R}\sin \psi,\;\varphi\approx {r\over R} \cos \psi.$$

Thus, a circle of radius $r$ in the coordinate system $(x, y),$ rotating with the apparatus, transfers into almost a circle of radius $r/R$ 
in a normalized polar coordinate system $(\rho/ R, \varphi)$ (see Figure 6).

\begin{center}
\includegraphics[height=1.7in]{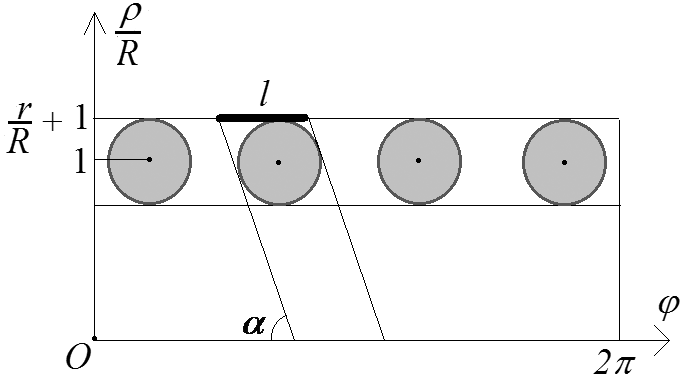}
\\{\bf Fig. 6.} 
The control area of one circle of radar scan in the normalized polar coordinate system  associated with moving vehicles.
\end{center}

When the asymptotic condition $r/R\ll 1$ is satisfied, the trajectory $A_{\psi}$ transforms into a straight line. If the linear speed of the vehicle is $v,$ and the speed of the mobile object in the fixed coordinate system is $u,$ then, in the polar coordinate system associated with the moving vehicles, the speed of the mobile object is approximately equal to $\sqrt{v^2/R^2+u^2/R^2}$ and makes an angle $\alpha=arctg {u\over v}$ with a horizontal line that is the image of circle $C_*.$ Therefore, each circle representing the control area of the radar locator overlaps a segment of length 
$l=\displaystyle{2n\over R\sin \alpha}$ for the mobile object (see Figure 6).

Let's move from the probability of the object detection to the minimum number of devices $M=\min\left(n:\;\displaystyle{ns(F_i)\over 2\pi}\geq 1\right),$ at which the probability of a mobile object detection is equal to one, then $M=\min\left(n:\;\displaystyle{nr\over \pi R \sin \alpha}\geq 1\right).$ If the number of devices is 
$n = M,$ then the distance between the centers of the neighboring survey circles can be obviously less than $l.$

Let $k_-<1<k_+,\;\max(|1-k_-|,\;|1-k_+|)\ll 1$ and a probability distribution $p (dk)$ on the segment $[k_-,k_+]$ is given such that $\displaystyle\int_{k_-}^{k_+}kp(dk)=1.$ We replace the radius $R$ on $kR.$ Then in the polar coordinate system, in virtue of the asymptotic condition $r/R\ll 1,$ an image of scan circle becomes approximately a circle of radius $\displaystyle{r\over kR}.$

\begin{center}\vskip -0.2cm 
\includegraphics[height=1.3in]{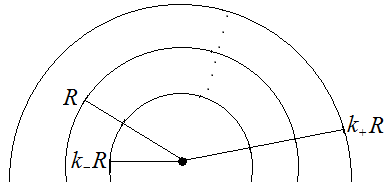}
\\{\bf Fig. 7.} 
The stochastization of the movement radius of the vehicles.
\end{center}

The function $\displaystyle{r\over kR}$ is convex downward by $k,$ hence, from Jensen's inequality \cite[p. 207]{Iensen}, we obtain the following inequality

\begin{equation}\label{Iensen}
{r\over R}\int_{k_-}^{k^+} {p(dk)\over k} \geq {r\over R} \cdot {1\over \int_{k_-}^{k^+} k p(dk)}={r\over R}.
\end{equation}

Thus, the probability of the object detection with a random distribution of the parameter $k$ increases in comparison with the probability of the object detection with an average value of the multiplier $k.$

\section{Possible generalizations of the model of the vehicles movement } 

Assume that vehicles move by variable courses with a linear velocity $v$ along a trajectory with a variable radius $k(t)R,\;k_-\leq k(t)\leq k_+,$ (see Fig. 8). This trajectory consists of segments along the radius and along the arcs of circles. Moreover, the total length of the arcs is small in comparison with the total length of segments along the radius. Let the process $k(t)$ obeys the hypothesis of ergodicity - $"$the ensemble average coincides with the trajectory average$"$, and has at $t\to\infty$ the limit distribution $p(dk),\;k_-\leq k\leq k_+.$ Then inequality (\ref{Iensen}) is fulfilled, which characterizes the probability of detection during the vehicles movement along a trajectory with a variable radius.

Now let the vehicles move along a segment $BA$ of length $R$ in one direction and, after reaching its end, turn and move in the opposite direction (see Fig. 9). We believe that the distance between neighboring vehicles is $2R / n,$ a similar model was considered in \cite{Kojemiakin}. 
\begin{center}\vskip -0.2cm
\includegraphics[height=1.3in]{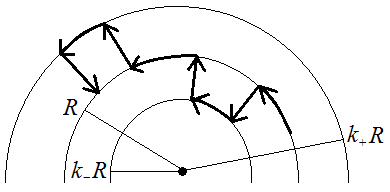}
\\{\bf Fig. 8.} The vehicle movement along a trajectory with a variable radius.
\end{center}

Then the scan circles associated with the vehicle locators move along the cylindrical surface $S$ with the generatrix $A_*A_{**}$ of length $2r$ and the directrix $B_*A_*B_*$ of length $2R.$ Strictly speaking, this surface is a strongly elongated elliptical cylinder (the mechanical analogue is a metal blank after complete deformation under the press).

\begin{center}\vskip -0.2cm
\includegraphics[height=1.35in]{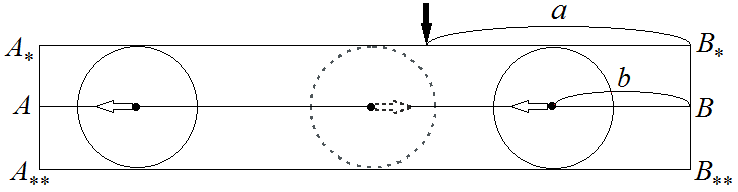}
\\{\bf Fig. 9.} 
Movements of the vehicles and the object in a fixed coordinate system (the bold arrow indicates the direction of the object movement).
\end{center}

In a fixed coordinate system, the object moves vertically, and its horizontal coordinate at the time of crossing the strip is  $a,\;0\leq a\leq R.$ 
Let transit to the coordinate system associated with the vehicles moving along the cylindrical surface $s,$ and denote as $b$ the distance from point $B$ to the nearest vehicle in the movement direction (from right to left in Figure 9).

Let the probabilistic distribution of the random vector $(a, b)$ be uniform on the rectangle $[0,R]\times[0,2 R/n].$ Then, in the coordinate system associated with moving vehicles, we can specify sub-segments of length $l=2r/\sin \alpha$ on the segment 
$A_*B_*$, where the object can be detected (see Figure 10).

The task of calculating the probability of the object detection is more complicated when the vehicles are moving along straight line, but not along a circle. 
\begin{center}
\includegraphics[height=1.1in]{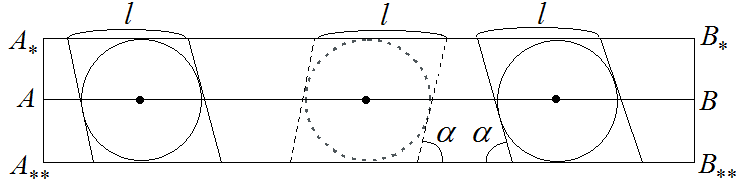}
\\{\bf Fig. 10.} 
The object movement in the coordinate system associated with moving vehicles.
\end{center}

A relatively simple solution to this problem was obtained by transition from the probability of the object detection to the minimum number of $M$ apparatuses, in which the probability of the object detection is equal to one, namely, $M=\min\left(n:\;\displaystyle {nr\over R\sin\alpha}\geq 1\right).$ Such a technique can be applied for multiagent systems, considered in \cite{Maggio}.

It should be emphasized in conclusion that we considered a dynamic version of the Buffon problem, which includes such characteristics as the trajectories and speeds of the mobile extraneous object and unmanned underwater vehicles. This circumstance leads to the necessity of transition to coordinate systems associated with moving vehicles and to additional, not always obvious, geometric constructions and application of the small parameter method.

The authors are grateful to corresponding member of Russian Academy Sciences A. F. Shcherbatyuk for valuable discussion of the paper.

Supported by Far Eastern Branch of Russian Academy Sciences, Program ``Far East`` (projects 18-5-050, 18-5-083).


\end{document}